# Latent Semantic Word Sense Disambiguation Using Global Co-occurrence Information

Minoru Sasaki

Department of Computer and Information Sciences, Faculty of Engineering,
Ibaraki University, 4-12-1, Nakanarusawa, Hitachi, Ibaraki, Japan
msasaki@mx.ibaraki.ac.jp

## ABSTRACT

*In this paper, I propose a novel word sense disambiguation method based on the global co-occurrence information using NMF. When I calculate the dependency relation matrix, the existing method tends to produce very sparse co-occurrence matrix from a small training set. Therefore, the NMF algorithm sometimes does not converge to desired solutions. To obtain a large number of co-occurrence relations, I propose to use co-occurrence frequencies of dependency relations between word features in the whole training set. This enables us to solve data sparseness problem and induce more effective latent features. To evaluate the efficiency of the method of word sense disambiguation, I make some experiments to compare with the result of the two baseline methods. The results of the experiments show this method is effective for word sense disambiguation in comparison with the all baseline methods. Moreover, the proposed method is effective for obtaining a stable effect by analyzing the global co-occurrence information.*

## 1. INTRODUCTION

In natural language processing, acquisition of sense examples from an example set that contain a given target word enables to construct an extensive data set of tagged examples to demonstrate a wide range of semantic analysis. For example, using the obtained data set, we can construct a classifier that identifies its word sense by analyzing co-occurrence statistics of a target word. Also, we can make a wide-coverage case frame dictionary automatically and construct thesaurus for each meaning of a polysemous word. To construct large-sized training data, language dictionary and thesaurus, it is increasingly important to further improve to select the most appropriate meaning of the ambiguous word.

If we have training data, word sense disambiguation (WSD) task reduces to a classification problem based on supervised learning. This approach is generally applicable to construct a classifier from a set of manually sense-tagged training data. Then, this classifier is used to identify the appropriate sense for new examples. A typical method for this approach is the classical bag-of-words (BOW) approach [5], where each document is represented as a feature vector counting the number of occurrences of different words as features. By using such features, it becomes easy to adapt many existing supervised learning methods such as Support Vector Machine (SVM) [1] for the WSD task. However, when the general vector space model, in which a document is represented as a vector using term frequency based weighting methods, is applied to the WSD, the local context words are typically used as features and the global context information without dictionary information is not employed in the previous research.

In this paper, I propose a novel word sense disambiguation method based on the global co-occurrence information using NMF. In previous research, [3] proposes a novel WSD method of particular word instances using the automatically extracted sense information. When we calculate the dependency relation matrix, the existing method tends to produce very sparse co-occurrence matrix from a small training set. Therefore, the NMF algorithm sometimes does not converge to desired solutions. To obtain a large number of co-occurrence relations, I propose to use co-occurrence frequencies of dependency relations between word features in the whole training set. This enables us to solve data sparseness problem and induce more effective latent features.

## 2. NON-NEGATIVE MATRIX FACTORIZATION

Non-Negative Matrix Factorization (NMF) is a popular decomposition method for multivariate data [4]. NMF decomposes the $m \times n$ non-negative matrix $X$ to the $m \times k$ matrix $W$ and the $k \times n$ matrix $H$, while these matrixes have no negative elements. Usually, $k$ is chosen to be smaller value than $n$ and $m$.

$$X \approx WH \quad (1)$$

Using the NMF for a term-document matrix $X$, the matrix $H$ represents the clustering result with $k$ topics.

For quantifying the quality of this approximation, cost functions based on Kullback-Leibler divergence is used and minimized using iterative update rules as follows:

$$W_{ij} \leftarrow W_{ij} \frac{(XH)_{ij}}{(WHH^T)_{ij}} \quad (2)$$

$$H_{ij} \leftarrow H_{ij} \frac{(X^T H)_{ij}}{(HW^T W)_{ij}}, \quad (3)$$

where $W_{ij}$ and $H_{ij}$ are the $i$-th row and the $j$-th column element respectively. These matrices $W$ and $H$ are initialized randomly with non-negative data and these above update rules are iteratively applied until the max iteration number (or convergence) is reached.

## 3. WSD USING GLOBAL CO-OCCURRENCE INFORMATION

### 3.1. Latent Semantic WSD Using Local Co-occurrence Information

In previous research, [3] proposes a novel WSD method of particular word instances using the automatically extracted sense information. This method induces latent features for three matrices. The first matrix $A$ contains co-occurrence frequencies of words that the target word co-occurs with. The second matrix $B$ contains term frequencies of words that appear in the context window. The third matrix $C$ contains co-occurrence frequencies of words that the co-occurring context words of the target word co-occur with. Then, NMF is applied to the three matrices to factorize each matrix into two non-negative matrices, while the former results are used to initialize the next factorization, as shown in Figure 1.

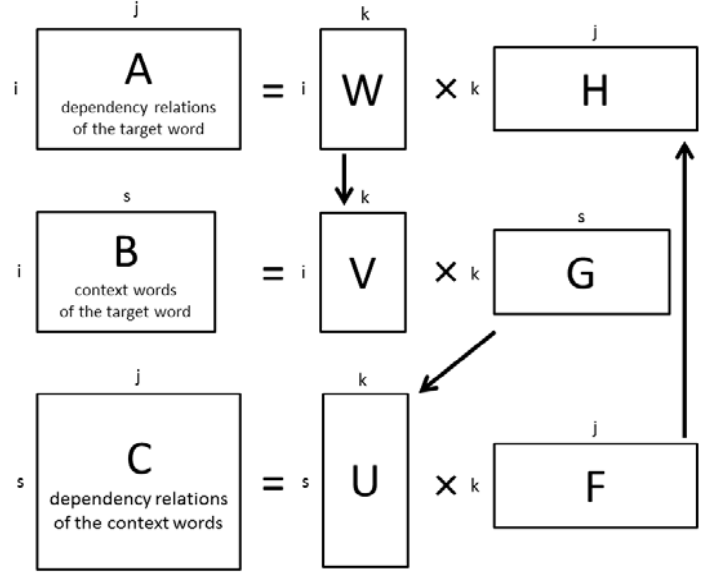

Figure 1. Interleaved NMF algorithm for Latent Semantic WSD

Given a non-negative matrices *A*, *B* and *C* in the beginning of this method, matrices *W*, *H*, *G* and *F* are initialized randomly with non-negative values. Then, it decomposes the matrix *A* into the matrices *W* and *H* using NMF. In the decomposition of the matrix *B*, the updated matrix *W* is copied to matrix *V* and the updated matrices *V* and *G* is computed using NMF. In the decomposition of the matrix *C*, the transpose of the updated matrix *G* is copied to matrix *U* and the updated matrices *U* and *F* are obtained using NMF. At the last step of the iteration, the matrix *F* is copied to matrix *H*. This iteration is repeated until the maximum number of iterations is reached or the objective function of all NMF decomposition no longer improves.

In order to perform this method for WSD, it needs to fold each sense of the target word into semantic space using the matrix *H*. For each sense in training data, centroid vector *c* is calculated and this centroid is mapped into the semantic space using the matrix *H* as follows:

$$b = cH^T \qquad (4)$$

For input example of the target word, its context words are extracted to construct a vector *f* and the vector *f* is also mapped into the same semantic space using the matrix *G* as follows:

$$d = fG^T \qquad (5)$$

Then, cosine similarity between the vector *d* and each of the sense vectors *b* are calculated and the sense that is the largest cosine similarity is selected.

### 3.2. Latent Semantic WSD Using Global Co-occurrence Information

This Latent Semantic WSD method is efficient for finding a reduced semantic space. However, problem arises when we apply this method. When we calculate the third matrix *C*, this method tends to produce very sparse co-occurrence matrix from a small training set. To obtain a large number of co-occurrence relations, I propose to use co-occurrence frequencies of dependency relations between word features in the whole training set. This enables us to solve data sparseness problem and induce more effective latent features. Like the above method, the proposed method needs three matrices, but the third matrix is different from the previous method. The third matrix *D* contains co-occurrence frequency of context words that co-occur in dependency relations to context words in a large document set. The proposed method induces latent features for these three matrices *A*, *B* and *D*.

## 4. EXPERIMENT

To evaluate the efficiency of the proposed WSD method using the global Co-occurrence information, I conduct some experiments to compare with the result of the existing methods. In this section, I describe an outline of the experiments.

### 4.1. Data

I used the Semeval-2010 Japanese WSD task data set, which includes 50 target words comprising 22 nouns, 23 verbs, and 5 adjectives [2]. In this data set, there are 50 training and 50 test instances for each target word. Moreover, to obtain a large number of co-occurrence relations, I use 22,832 documents chosen from the Japanese BCCWJ corpus[1].

### 4.2. Evaluation Method

#### 4.2.1. Baseline System 1

As the first baseline method, I only use the first matrix *A* described in section 3.1. To construct the matrix *A*, I represent each sentence with the target word in the training set as a high-dimensional vector where each component represents the co-occurrence frequency of the target word in the sentence. Then, NMF is applied to the matrix *A* to factorize each matrix into two non-negative matrices *W* and *H*. Each vector is tagged with the sense of the target word in that sentence. So centroid $c_i$ of the co-occurrence vectors that are assigned the same sense $i$ is calculated and each centroid $c_i$ is mapped into the semantic space using the matrix *H* as follows:

$$b_i = c_i H^T \qquad (6)$$

For input example of the target word, its context words are extracted to construct a vector *f* and the vector *f* is also mapped into the same semantic space using the matrix *H* as follows:

$$d = f H^T \qquad (7)$$

Then, cosine similarity between the vector *d* and each of the sense vectors *b* are calculated and the sense that is the largest cosine similarity is selected.

#### 4.2.2. Baseline System 2

In the second baseline system, I use the latent semantic WSD using local co-occurrence information described in Section 3.1. I construct the three matrices *A*, *B* and *C* to induce latent semantic dimensions using NMF.

Table 1. Experimental Results of Each Execution
(highest precision rates among all the run are written in bold font)

| System | Run 1 | Run 2 | Run 3 |
|---|---|---|---|
| Baseline System 1 | 53.28% | **53.88%** | 51.80% |
| Baseline System 2 | 59.68% | **61.08%** | 61.08% |
| Proposed Method | 60.44% | **61.48%** | 61.04% |

---

[1] http://www.ninjal.ac.jp/english/products/bccwj/

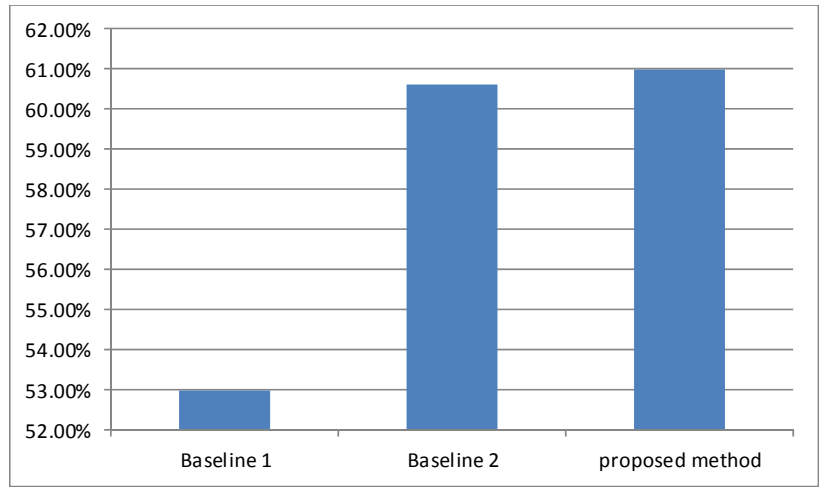

Figure 2. Highest Precision of Each system

## 5. EXPERIMENTAL RESULTS

Figure 2 shows the experimental results of the baseline methods and the proposed method. Computational experience reported shows that the choice of initial point is quite important to the NMF's goal. In practice, the algorithms are run several times with different initial points and the NMF is chosen as the feasible solution. In our experiments, each method is executed three times and average precision of all execution is calculated.

In this Figure 2, the proposed method shows higher precision than the other baseline methods so that this approach is effective for word sense disambiguation. In comparison with the baseline system 1, the proposed method can obtain better precision so that it is effective for WSD to use context information and co-occurrence information. In comparison with the baseline system 2, the proposed method provides slightly better precision than the baseline system 2. As shown in Table 1, the proposed method can obtain the highest precision and can be stable at high precision value. However, the baseline system 2 cannot achieve stable precision because of the lack of the number of co-occurrence information. Therefore, the proposed method is effective for obtaining a stable effect by analyzing the global co-occurrence information.

## 6. CONCLUSION

In this paper, I propose a novel word sense disambiguation method based on the global co-occurrence information using NMF. To evaluate the efficiency of the method of word sense disambiguation, I conduct some experiments to compare with the result of the two baseline methods. The results of the experiments show this method is effective for word sense disambiguation in comparison with the all baseline methods. Moreover, the proposed method is effective for obtaining a stable effect by analyzing the global co-occurrence information.

Further work would be required to consider a larger sized training data to obtain a large amount of co-occurrence information.

**Authors**

Minoru Sasaki: received his B.E., M.E. and D.Eng. degrees in information science and intelligent systems from the University of Tokushima in 1996, 1998 and 2000. He is now a lecturer in the department of computer and information sciences at Ibaraki University. His research interests include natural language processing and information, information retrieval and text mining.

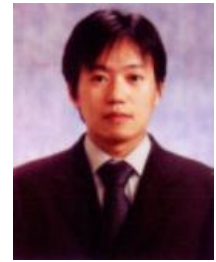